# On the evolution of research in hypersonics: application of natural language processing and machine learning

Ashkan Ebadi[1,*], Alain Auger[2], and Yvan Gauthier[3]

[1] Digital Technologies, National Research Council Canada, Montreal, Quebec, Canada
[2] Science and Technology Foresight and Risk Assessment Unit, Defence Research and Development Canada, Ottawa, Ontario, Canada
[3] Digital Technologies, National Research Council Canada, Ottawa, Ontario, Canada
[*] Correspondence: ashkan.ebadi@nrc-cnrc.gc.ca

**Abstract**

Research and development in hypersonics have progressed significantly in recent years, with various military and commercial applications being demonstrated increasingly. Public and private organizations in several countries have been investing in hypersonics, with the aim to overtake their competitors and secure/improve strategic advantage and deterrence. For these organizations, being able to identify emerging technologies in a timely and reliable manner is paramount. Recent advances in information technology have made it possible to analyze large amounts of data, extract hidden patterns, and provide decision-makers with new insights. In this study, we focus on scientific publications about hypersonics within the period of 2000-2020, and employ natural language processing and machine learning to characterize the research landscape by identifying 12 key latent research themes and analyzing their temporal evolution. Our publication similarity analysis revealed patterns that are indicative of cycles during two decades of research. The study offers a comprehensive analysis of the research field and the fact that the research themes are algorithmically extracted removes subjectivity from the exercise and enables consistent comparisons between topics and between time intervals.

**Introduction**

The term "hypersonic" is used to characterize objects that move faster than 5 times the speed of sound (Definitions 2022). Hypersonic technology is not a new research topic and has been investigated for more than six decades (Blankson and Pyle 1993; Czysz and Vandenkerckhove 2000), although international interest varied over the years (Van Wie 2021). With applications in several domains, including military (Malinowski 2020) and commercial (Ingenito et al. 2011), hypersonic technology is argued to be crucial for both national defence and space exploration purposes (Gu and Olivier 2020). Given the competition between states, hypersonics is now of increased importance (Van Wie 2021), and it has become a hot research topic in the scientific community as well as in international relations.

Over sixty years of hypersonics technology investigation has resulted in both experimental and operational systems (Van Wie 2021). High-speed vehicles have attracted a growing interest in the aviation field, supporting manned and unmanned operations/explorations in low earth orbit (LEO), calling for high flexibility, affordability, and high safety standards (Viviani and Pezzella 2019). In addition to space exploration missions, interest has been recently growing in hypersonics travels for commercial/civilian application, with many start-ups focusing on hypersonic aircrafts (Viviani and Pezzella 2019). Therefore, many researchers in basic and applied technologies have focused on hypersonic research and development due to the potential advantages of an operational hypersonic vehicle (HV) (Viviani and Pezzella 2019).

There are several examples of activities/initiatives that have contributed to engineering advancement, for example: 1) the X-15 research aircraft (1959-1968), 2) the Apollo reentry capsule (1966-1975), 3) the Space Shuttle program (1981-2011), and 4) the SpaceX Falcon 9 launch system (2010-present) (Van Wie 2021). From a military perspective, aviation and space have been always critical for nations from a national defence perspective (Richman et al. 2019). As a result of this global attention, new categories of hypersonic capabilities have emerged or are being explored. This includes, but is not limited to, hypersonic boost-glide systems, interceptor missiles, reusable aircrafts, hypersonic cruise missiles (HCM), and gun-launched projectiles (Van Wie 2021). Technology emergence is particularly noticeable in the area of offensive hypersonic strike systems, followed by the emergence of hypersonic defence capabilities to counter these offensive systems (Van Wie 2021). To date, the United States, China, and Russia have the most advanced hypersonic military capabilities (Richman et al. 2021).

With efforts in hypersonics research accelerating, applications of the technology in different markets are becoming operational (Deloitte 2020). Beyond military applications, hypersonic technology has potential for commercial applications, including reusable hypersonic systems/aircrafts and fully reusable space-access vehicles (Van Wie 2021). For instance, reusable hypersonic aircrafts are expected to be developed for routine commercial flights (Carioscia et al. 2019). Such vehicles will require additional technology advancement to address issues inherent to interacting components such as propulsion, structures, and guidance and control systems (Van Wie 2021).

Governments are investing in hypersonics technology as part of their national defence strategies. In the United States only, more than $2.6 billion is annually invested, with a 26% compound annual growth rate (CAGR) since 2014 (Deloitte 2020). The international hypersonic market is predicted to develop at a CAGR of 9.5% during the period of 2022 to 2028, with the global market value exceeding $12.5 billion by 2028 (Facts and Factors 2020). In addition, venture capital investment in hypersonic technologies is expected to grow in the coming years in response to market demand (Deloitte 2020). The research activities have also been fueled by significant investments. For instance, the United States has made a more than 7-fold funding increase in hypersonic research between 2015 and 2020 (McDonald 2021). The re-usability and speed of hypersonic vehicles have been key driving factors of these investments (Piplica 2021), especially on the commercial side, and hypersonic air travel is now expected to become operational within the decade (McDonald 2021).

For the scientific community, peer-reviewed scientific publications represent the main channel to communicate discoveries and findings. Due to the availability of large-scale publication data, quantitative analysis of such data has been widely carried out to extract insights from the corpus of publications and the relationships between authors and articles (Ebadi et al. 2020). Recent advancements in the field of computer science provide a unique opportunity to go beyond traditional methods, explore massive high-dimensional data sets, and obtain new insights. For example, topic modeling (TM) is an unsupervised machine learning technique that can automatically organize a collection of documents, in our case scientific publications, into a set of high-level themes, so-called topics (Papadimitriou et al. 2000), leveraging context clues to infer hidden themes in the unstructured text data. TM has been widely used for extracting latent topics from the collection of documents in various domains, e.g., psychology (Bittermann and Fischer 2018), education and information technologies (Ozyurt and Ayaz 2022), transportation research (Das et al. 2016), and the recent COVID-19 pandemic (Ebadi et al. 2021), to name a few.

Considering the economic and national defence implications of hypersonic technologies as well as their fast-evolving nature, it is crucial for policy- and decision-makers to have a comprehensive understanding of the related R&D landscape. However, this often translates into a manual investigation of a huge set of data sources and/or documents, which apart from being time-consuming and not scalable, may also suffer from a lack of comprehensiveness, biases, and subjectivity. The high availability of digitalized research data on one hand, and recent advances in analytics and machine learning research, on the other hand, present new opportunities for large-scale advanced data analytics. These advanced approaches and tools can assist human analysts by revealing hidden patterns and new insights from massive data sets (Keller and Heiko 2014).

The main objective of this study is to obtain a better understanding of the hypersonics research landscape and its evolution over time. To that end, we leverage natural language processing (NLP) and machine learning to analyze quantitatively the scientific publications about hypersonics within the period from 2000 to 2020. To the best of our knowledge, this is the first study that performs a comprehensive analysis on hypersonics publications to characterize its research landscape and evolution. Our proposed approach can provide policy-makers and strategic planners with a high-level representation of the hypersonics technology landscape, as well as the main research themes, their relationships, and temporal progression, all automatically being extracted from a large corpus.

The remainder of this paper is as follows. Section "Data and methodology" describes data and techniques in detail. Section "Results" presents the findings of the research. Findings are then discussed and the conclusions are presented in Section "Discussion and conclusion". Finally, some limitations of the research and future directions are presented in Section "Limitations and future work".

## Data and methodology

### Data

The scope of this research covers all research publications about hypersonics that are accessible through Elsevier's Scopus, i.e., a comprehensive abstract and citation database of peer-reviewed scientific journals, books, and conference proceedings. Initially, the bibliographic data of hypersonics-related publications, published within the period of 2000 to 2020, were retrieved from Elsevier's Scopus by running "hypersonic*" as the search query[1]. The collected bibliographic data included meta-data about each of the extracted papers, such as the title, abstract, date of publication, list of authors, and their affiliations. We filtered out publications for which neither title nor abstract was available. This initially resulted in a total of 17,075 publications on which we applied several preprocessing steps that will be explained in detail in Section "Text pre-processing".

### Methodology

The methodology has five main components: 1) text pre-processing, 2) correspondence analysis, 3) lexical complexity analysis, 4) publications similarity analysis, and 5) structural topic modeling. These components are introduced and discussed in detail below.

---

[1] Data were collected in August 2021.

*Text pre-processing*

The abstract of a publication provides a condensed representation of its content. Titles may also provide some complementary informative keywords and/or keyphrases that may not be necessarily present in the abstract section. Since we did not have access to the entire texts of the publications, we decided to integrate for each paper both the title and abstract (and not just use one of them) to obtain a better representation of the publications' content. As the first pre-processing step, we created a new column in the dataset by combining the title and abstract of each publication and removed duplicated records. Next, we applied several pre-processing steps to prepare the data for analysis. These steps involved converting text to lowercase, removing stop words based on a customized English stop words list, correcting special characters, and removing punctuations. We tokenized the processed data and created a document-term frequency matrix in which each row represents one publication, columns represent the tokenized terms, and each cell value is the number of appearances of a given term in a given publication. The final dataset contains 17,075 publication records.

*Correspondence analysis*

Correspondence factor analysis (Greenacre and Blasius 2006) was first applied to the pre-processed data (the document-term frequency matrix) to verify the existence of a temporal trend in publications. Correspondence analysis (CA) provides a graphical representation of cross-tabulations by mapping data on visually understandable dimensions in which noises are filtered out, making it easier to visualize general patterns in the data (Blasius and Greenacre 1998; Doré and Ojasoo 2001). Using the generated term-frequency matrix (as explained in Section "Text pre-processing"), we created a term-frequency matrix for each year of the examined period by filtering on the year of publication. Next, terms with a frequency less than 50 in each year were filtered out in order to reduce/remove noise. Highly frequent terms with appearance in more than 60% of the publications in each year were also removed as they were common terms and not specific enough to provide insight on the general temporal pattern. The filtered term-frequency matrix was then mapped to a 2-dimension (2D) space by extracting the first two principal components. Results are discussed in Section "Temporal trend verification".

*Lexical complexity analysis*

As a preliminary step in understanding the evolution of the scientific terminology in the examined case technology, we investigated the linguistic complexity of the publications. Specifically, using abstracts of publications in the dataset, we analyzed their textual complexity from two aspects, i.e., textual readability and richness.

To assess textual readability, we calculated the following 5 measures per paper and averaged over each year of the examined time interval: 1) number of words per paper, 2) the number of sentences, 3) sentence length, defined as the number of words divided by the number of sentences, 4) ratio of difficult words, defined as words that are not common and has at least 2 syllables, and 5) the Flesch–Kincaid score (Kincaid et al. 1975) that indicates how difficult a text is to understand using sentences, words, and syllables as the core components. Lower Flesch–Kincaid scores demonstrate texts that are easier to read while higher numbers indicate more difficult texts.

The following 3 measures were calculated to assess lexical richness: 1) type-token ratio (TTR), 2) measure of textual lexical diversity (MTLD), and 3) hypergeometric distribution diversity (HD-D). As the most well-known measure of lexical diversity (Lissón and Ballier 2018) and as a measure of vocabulary variation, TTR is defined as the ratio of the total number of unique words, i.e., types, divided by the total number of words, i.e., tokens, in a given text. The MTLD measure, developed by McCarthy (2008), divides the text into segments that are of variable lengths since fragmentation is done based on the TTR values of the segments, until it reaches the default TTR size value (0.72). The HD-D measure (McCarthy and Jarvis 2007) is based on the hypergeometric distribution representing the probability of finding a certain number of words from a random sample of a certain size. Results are discussed in Section "[Lexical complexity analysis](Lexical complexity analysis)".

*Publications similarity over time*

After verifying the existence of a temporal trend in the corpus (cf. "[Correspondence analysis](Correspondence analysis)"), we investigated research similarities. This preliminary analysis would shed light on the evolution rate of the research terminology in the examined case technology, i.e., hypersonics. For this purpose, a Doc2Vec model (Le and Mikolov 2014) was trained on the corpus to learn publication-level embeddings. Finally, cosine similarity was applied to the publications' embedding vectors, and results were aggregated for each year of the examined period to evaluate publications' similarity over time. Results are discussed in Section "[Publications similarity over time](Publications similarity over time)".

*Structural topic modeling*

To complement the previous steps, topic modeling (TM) was applied to extract latent research themes from the corpus. Unlike conventional methods (e.g., content analysis (Krippendorff 2018)) that require considerable manual effort (Gatti et al. 2015), as an unsupervised machine learning technique, TM can summarize huge text data collections and extract latent semantic themes automatically (Blei et al. 2003). In other words, TM is an approach to cluster few words across the corpus into topics (Bhat et al. 2020). Topic modeling has been widely used in the literature to analyze the research landscape of various domains, including but not limited to transportation (Sun and Yin 2017), cancer (Mosallaie et al. 2021; Stout et al. 2018), manufacturing (Yoon et al. 2019), and even the recent COVID-19 pandemic (Ebadi et al. 2021).

In this work, we used Structural Topic Modeling (STM) (Roberts et al. 2019) since it has two main properties that were critical for our research objectives: 1) it allows incorporating publication-level covariates of interest for which we considered publication date as a covariate to analyze the temporal evolution of the domain, and 2) it allows topics to be correlated which helped us to better understand the dynamics of the hypersonics research as well as the structure of the topics at the corpus level. The STM model was built on the entire dataset with an annual granularity to capture hidden temporal patterns necessary to draw the landscape of hypersonics research and analyze its evolution. Similar to other TM approaches, STM requires no data labeling and the topics emerge automatically in an unsupervised setting (Ebadi et al. 2020).

STM requires the number of topics to be set as a fixed parameter. There is no universal quantitative approach for finding the optimal number of topics in topic models (Lucas et al. 2015). Since a fully automated approach to finding the optimal number of topics may result in inaccurate findings (Maskeri et al. 2008), we used a multi-layer semi-automated approach to determine the number of topics. For this purpose, several baseline Latent Dirichlet Allocation (LDA) models (Blei et al. 2003) were first built by varying the number of topics in the [3, 20] range and the coherence of the generated topics was calculated. We used the $C_v$ score (Röder et al. 2015), which is an intrinsic evaluation metric, for topic coherence calculation. The $C_v$ scores helped us to narrow down the optimal range for the number of topics further. It was found that the optimal number of topics should be in the range of [10, 15]. Next, we manually checked the top keywords and keyphrases assigned to each topic in the generated models and analyzed topic-word distributions.

Based on this manual topic verification step, it was concluded that the optimal number of topics for the examined corpus is 12 and hence, the STM model was built with 12 topics. Since STM does not assign a representative label to the extracted topics automatically and in order to ensure the quality of the model, the extracted topics were finally reviewed by senior scientists with expertise in hypersonics research from Defence Research and Development Canada. They assigned a meaningful and concrete label to the topics. Figure 1 shows the conceptual flow of the study.

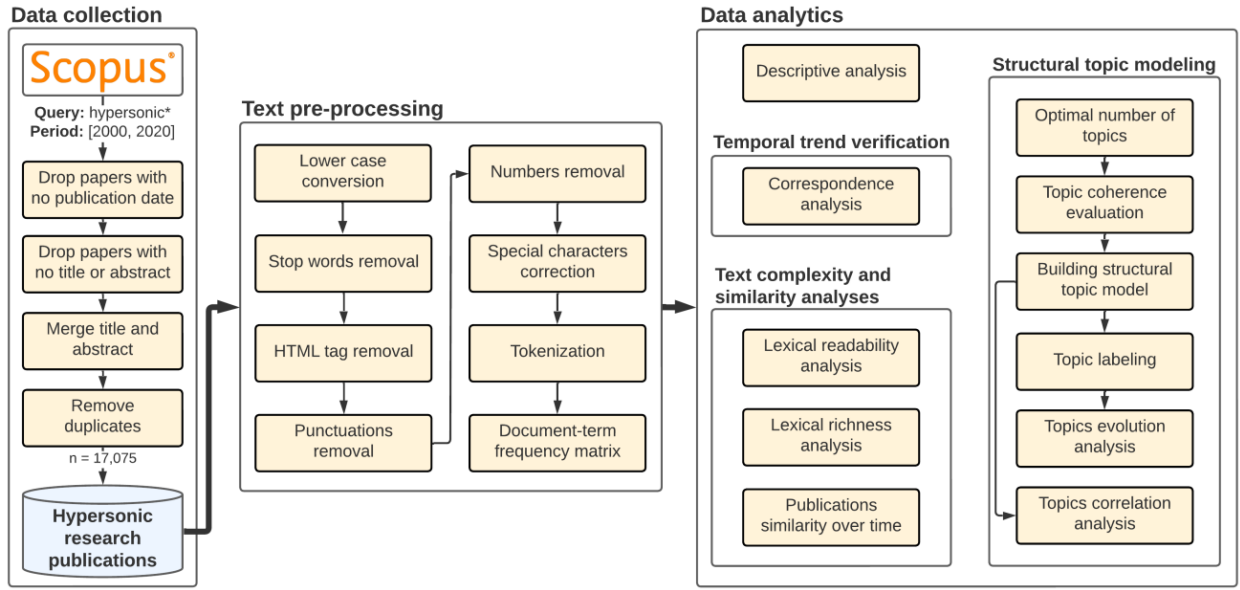

**Figure 1.** The conceptual flow of the analyses. After data collection, the pipeline contains five main steps, i.e., text pre-processing, temporal trend verification, lexical complexity analysis, publications similarity analysis, and structural topic modeling. In the data collection step, hypersonics publications within the period of [2000, 2020] are collected. Data is then filtered and pre-processed and passed to the data analytics. After verifying the existence of temporal evolution by correspondence analysis, lexical complexity is investigated, and publications similarities are analyzed over time. Finally, a structural topic model is built to extract the main research themes, their keyword sets, as well as the temporal trends.

## Results

### Number of publications trend

Figure 2 shows hypersonics publications distribution over the examined period. We marked an article as a Canadian publication if at least one of the authors had a Canadian affiliation. Canadian publications are color-coded in red in Figure 2. The number of publications follows an increasing trend, and almost tripled between the beginning and the end of the period. Canadian publications maintained a constant output since 2008, except for a drop in 2016. The overall publication rate plateaued between 2017 and 2020. The observed growth in the number of publications is in line with several studies (e.g., Bornmann and Mutz 2015; Ebadi et al. 2020; Zeng et al. 2019) that reported such an increasing trend in multiple domains. One may note that the growth could be also partially due to higher coverage of publications in Scopus during the final years of the period under review.

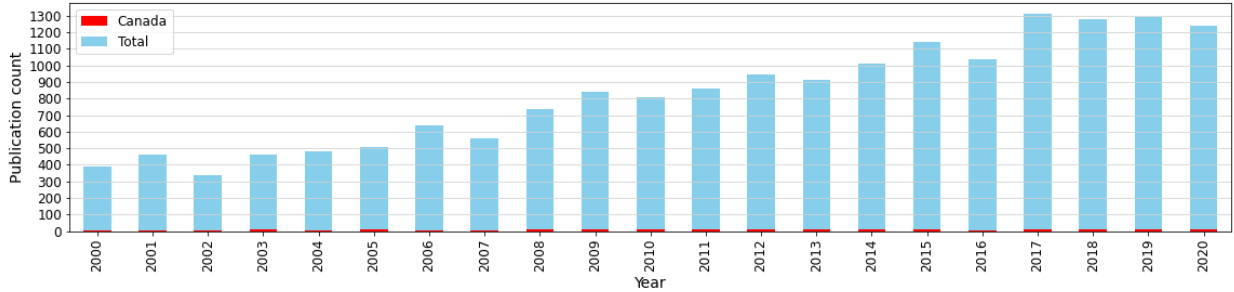

**Figure 2.** The trend of hypersonics publications between 2000 and 2020. Blue and red bars represent the total number of publications and publications by at least one Canadian author, respectively.

**Temporal trend verification**

Before analyzing the temporal evolution of the hypersonics research themes, the existence of the temporal trend was verified by performing a correspondence analysis on the frequent terms extracted from publications in each year of the examined time interval (as explained in Section "Correspondence analysis"). As seen in Figure 3, the research terminology has evolved over time, forming a relatively U-shape curve. Based on the distance of the publication date variables, i.e., the orange points in the figure, from the axes and the origin, it is clear that the frequent terms have evolved over time which confirms the existence of a temporal trend in the examined corpus.

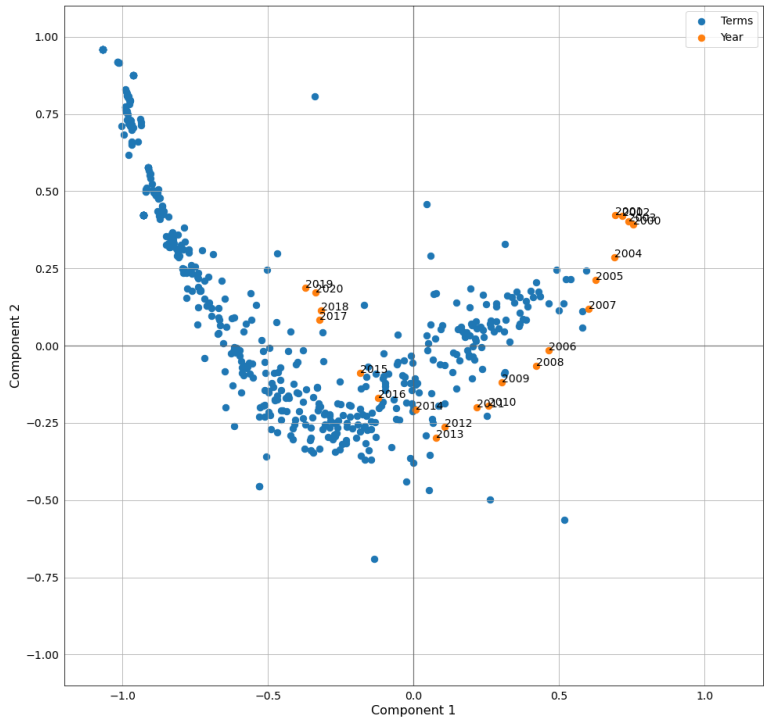

**Figure 3**. Correspondence analysis on the frequent terms appeared in the hypersonics publications from 2000 to 2020. Years of publications and their representative frequent terms are depicted with orange and blue points respectively.

**Lexical complexity analysis**

Figure 4 shows the results for the readability measures. From Figure 4-a, it is observed that the average number of words in the abstract section of the publication has followed an increasing trend after 2010, after some fluctuations at the beginning of the examined time interval, and reaching its peak in 2020. The number of sentences (Figure 4-b) has increased sharply after 2013, while experiencing a smoother increasing trend beforehand. However, as observed in Figure 4-c, the average length of the sentences has decreased. From Figure 4-a, -b, and -c, it can be said that researchers tend to use more sentences of shorter lengths recently that contain relatively more common (easy) words. Figure 4-d shows the trend of the ratio of the difficult words used in the publications' abstract. The difficult word ratio has also a decreasing trend with a steeper slope after 2013. The last sub-figure, i.e., Figure 4-e, depicts the Flesch-Kincaid score. Intuitively, this score reflects the number of years of education that is required to understand the text. Figure 4-d and -e show an increase in lexical readability over time, indicating that researchers have gradually used fewer difficult words in the abstract section to (maybe) engage a broader audience.

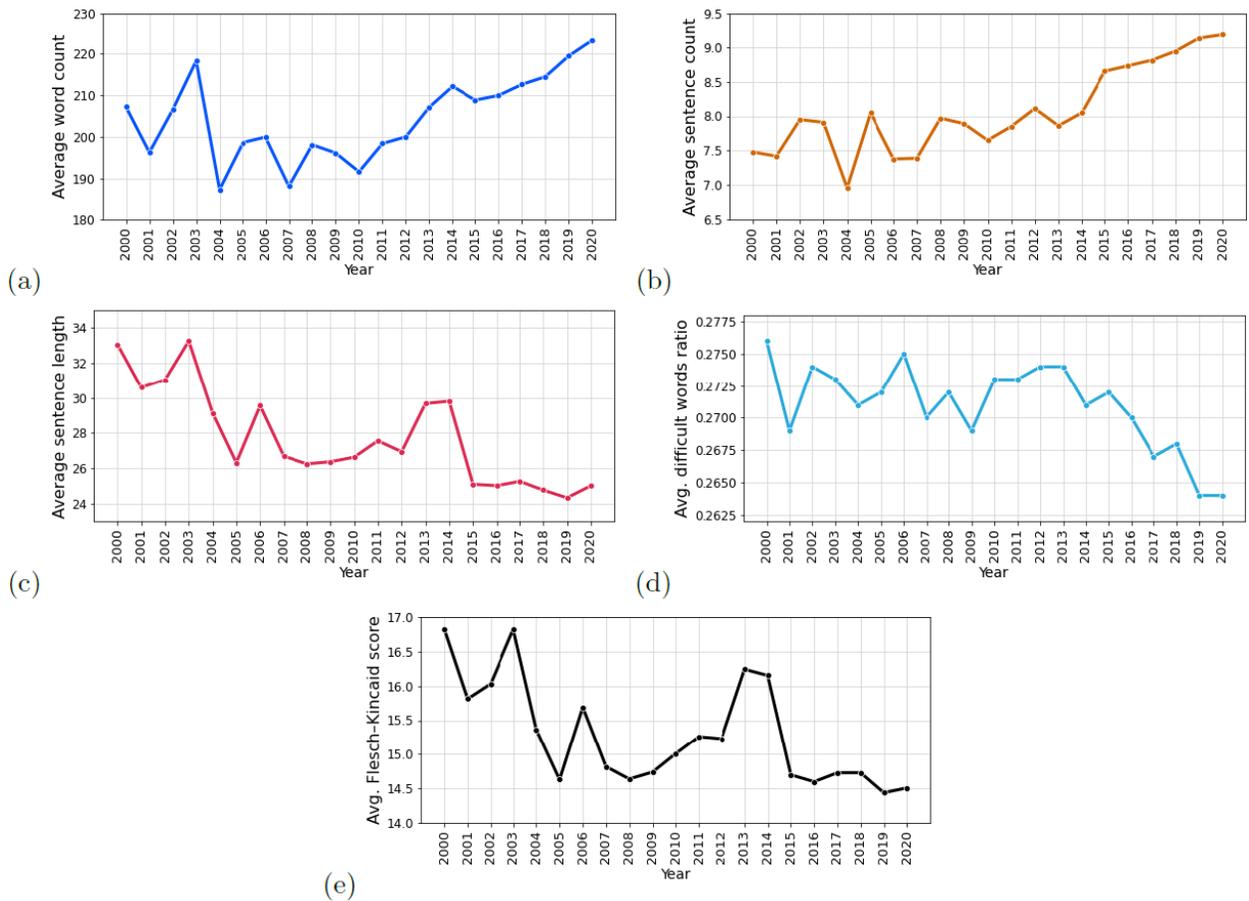

**Figure 4.** Lexical readability, **(a)** average word count, **(b)** average sentence count, **(c)** average sentence length, **(d)** average ratio of difficult words, and **(e)** average Flesch-Kincaid score.

Figure 5 shows measures of lexical richness. As seen in Figure 5-a, after a relatively constant trend in the beginning years, the average type-token ratio (TTR) has followed a decreasing trend, indicating that the vocabulary in hypersonics research publications has become less varied, and more uniform. One problem reported for TTR is that it does not generalize well for long texts (Johansson 2008), however, this is not the case here as we performed the analysis only on the abstracts of publications. Figure 5-b shows the average hypergeometric distribution diversity (HD-D) score. The intuition behind the HD-D score is that if a sample contains many forms of a specific word, drawing a sample containing at least one form of that word would be very likely (Fergadiotis et al. 2013). The trend of the measure of textual lexical diversity (MTLD) is depicted in Figure 5-c. HD-D and MTLD results also confirm that the diversity of the research vocabulary has decreased over time. This could be an indicator that the field and its terminology are stabilizing, however; further investigation is required to confirm this observation. HD-D value increases with text length while MTLD value decreases with text length (Treffers-Daller et al. 2018). The fact that both mentioned measures indicate a decreasing trend for lexical richness also verifies that text length is not an influencing factor in our performed analysis.

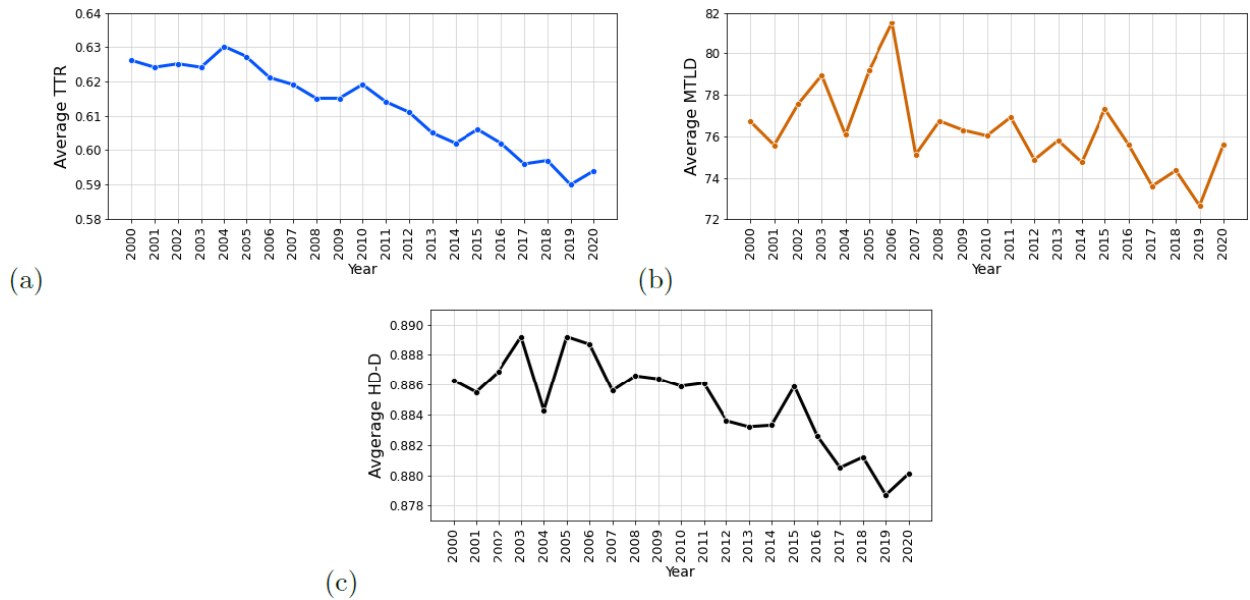

**Figure 5.** Lexical richness, **(a)** average type-token ratio (TTR), **(b)** average hypergeometric distribution diversity (HD-D) score, and **(c)** average measure of textual lexical diversity (MTLD).

**Publications similarity over time**

Figure 6-a shows the between-years similarity of hypersonics publications. Papers published in 2020 have on average the highest similarity with the publications in the previous years. For publications in 2020, the highest similarity is observed with publications in 2000, 2002, and 2004. This reference to earlier scientific findings may partially indicate modification of earlier results and/or reinterpretation of theories and conceptual frameworks. The other observation is that publications in 2001 have the lowest similarity with the other years in the examined time interval which could indicate the specificness and/or exclusivity of the scientific ideas that were investigated in that year. Figure 6-b depicts research similarities in consecutive years. As observed, the similarity score is relatively high, fluctuating in the range of 0.60 and 0.65. In addition, the similarity between hypersonics publications has followed an increasing trend. These observations may partially shed light on the evolution of hypersonics research and its cycles. We further investigate the evolution of the examined field in the next section.

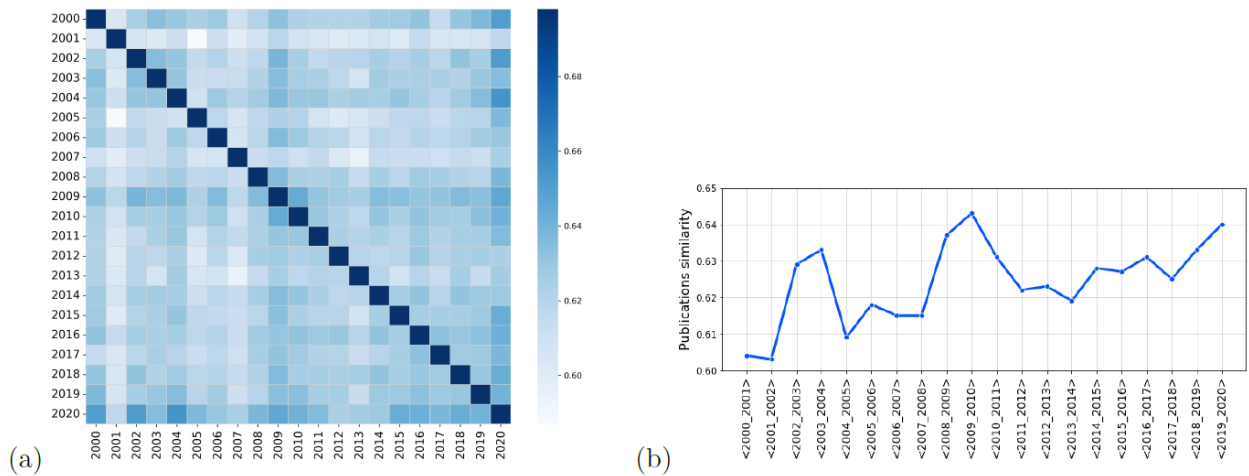

**Figure 6.** Publications similarity, **(a)** between years research similarity heatmap, and **(b)** consecutive years similarity.

**Hypersonics research themes and their temporal evolution**

In this section, we first extract the main research themes in the hypersonics field and analyze their correlation and quality. Then, we analyze the temporal evolution of the research themes in more detail.

*Hypersonics research themes*

As explained in Section "Structural topic modeling", the structural topic modeling (STM) technique (Roberts et al. 2019) was used to extract 12 main research themes in the field of hypersonics, using the year of publication as the covariate. The respective keyword sets of the extracted research themes were carefully reviewed and verified by senior scientists from Defence Research and Development Canada with domain expertise in hypersonics, and a representative label was assigned to each of the extracted themes as follows, terms in parentheses represent the respective high-level category:

- **Topic 1:** Cooling materials (*aerothermodynamics*)
- **Topic 2:** Fuel flow/propulsion systems (*aerothermodynamics*)
- **Topic 3:** Flight systems design (*aerothermodynamics*)
- **Topic 4:** Physico-chemical modeling (*aerothermodynamics*)
- **Topic 5:** Hypersonic velocity (*materials and structures*)
- **Topic 6:** Shock wave/boundary layer interaction-SWBLI (*materials and structures*)
- **Topic 7:** Flight vehicles tracking control (*guidance, navigation and control*)
- **Topic 8:** Wing panel modeling (*modeling, simulation and analysis*)
- **Topic 9:** Wind tunnel testing (\*modeling, simulation and analysis*)
- **Topic 10:** Hypersonic materials (*materials and structures*)
- **Topic 11:** Hypersonic inlets (*vehicles, propulsion and fuels*)
- **Topic 12:** Guidance against hypersonic targets (*guidance, navigation and control*)

As seen in the above list, 4 research themes belong to the *aerothermodynamics*, 3 belong to the *materials and structures*, 2 to *guidance, navigation and control* and *modeling, simulation and analysis*, and 1 to the *vehicles, propulsion and fuels* categories, respectively. In extracting topics, we followed the approach proposed in (Bischof and Airoldi 2012) to improve the quality of the

keywords, by extracting keywords that were both frequent and exclusive. Figure 7 shows the extracted topics' quality assessed through a combination of semantic coherence of the keywords of a given topic and exclusivity of the keywords to the topic. The size of the markers in the figure represents the proportion of the publications covered by the topic. Semantic coherence is calculated based on the number of times the keywords of a topic co-occur in publications. By this definition, a topic with a higher semantic coherence score could be considered as being more interpretable to humans. On the other hand, a topic is exclusive if keywords with a high co-occurrence likelihood conditioned on the topic have a low likelihood conditioned on other topics (Kuhn 2018). As seen in the figure, the extracted topics have high semantic coherence and exclusivity and most of them are located in the upper right region of the plot. Among the extracted topics, *Topic 1*, i.e., cooling materials, and *Topic 4*, i.e., physico-chemical modeling, have the highest exclusivity and coherence, respectively.

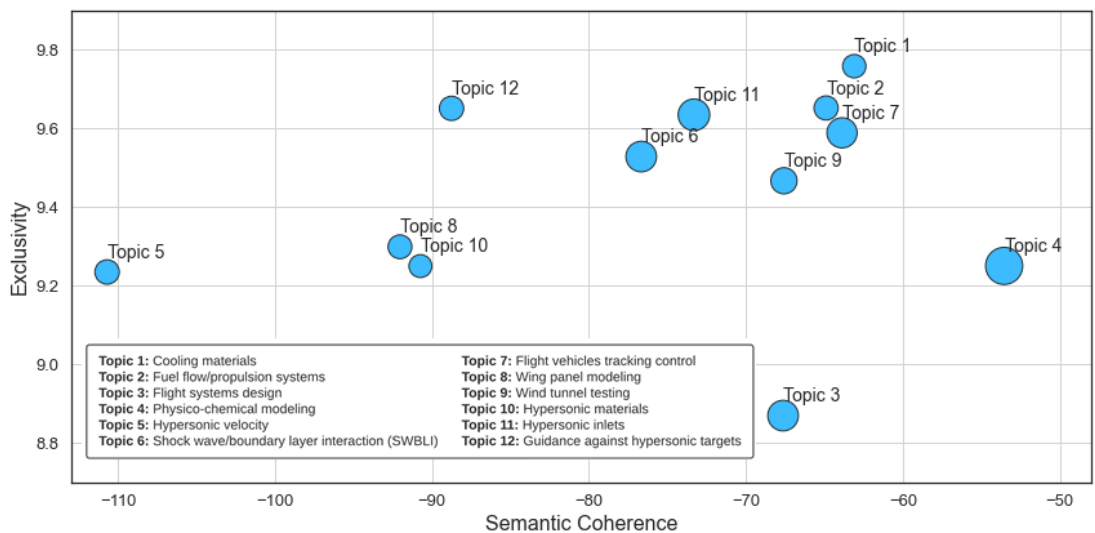

**Figure 7.** Quality of the extracted research themes based on semantic coherence and exclusivity of their representative keywords. Topic markers are sized based on the number of publications covered by the topic.

It should be noted that these extracted research themes only represent the main research areas of interest in the field of hypersonics at an abstract level, based on the collected scientific publications, and by no means do they capture all the details in the examined research area. STM allows topics to be correlated. To analyze interrelationships between the extracted topics and indicate topics that are likely to co-occur within the same publication, topic correlation analysis was performed. As seen in Figure 8-a, most of the extracted topics either do not correlate or are negatively correlated meaning that it is very unlikely for them to occur in the same paper. Figure 8-b shows the topics correlation graph in which two nodes, i.e., topics, are connected if they are positively correlated. The thickness of an edge linking two nodes represents the correlation coefficient. As observed, the correlation coefficient is not high. The four topic pairs that are positively correlated in descending order are: <Topic 12, Topic 7>, <Topic 5, Topic 10>, <Topic 1, Topic 10>, and <Topic 6, Topic 9>. One may note that these correlations might have been affected by journal coverage and indexing over time. However, for instance, a positive correlation between topic 12, i.e., guidance against hypersonic targets, and topic 7, i.e., tracking control of hypersonic flight vehicles is reflected in the literature as well in, e.g., guidance and control design of interceptors (Song and Zhang 2015), guidance algorithms for hypersonic reentry vehicles (Zang et al. 2019), and guidance and control for hypersonic missiles (Liu et al. 2017), to name a few. The uncertainties in hypersonic flight vehicles (HFVs) dynamics due to their specific characteristics such as fast time-varying flight environment and strong dynamics coupling (Li et al. 2020), as well as non-linearities enhancement as a result of the integration of the propulsion system and the body, have made HFVs' controller design a complicated task (Bu et al. 2015) that requires expertise in several sub-fields. The topic correlation analysis indicates the existence of more distinct research areas, despite the observed diversity and overlaps between a few topics.

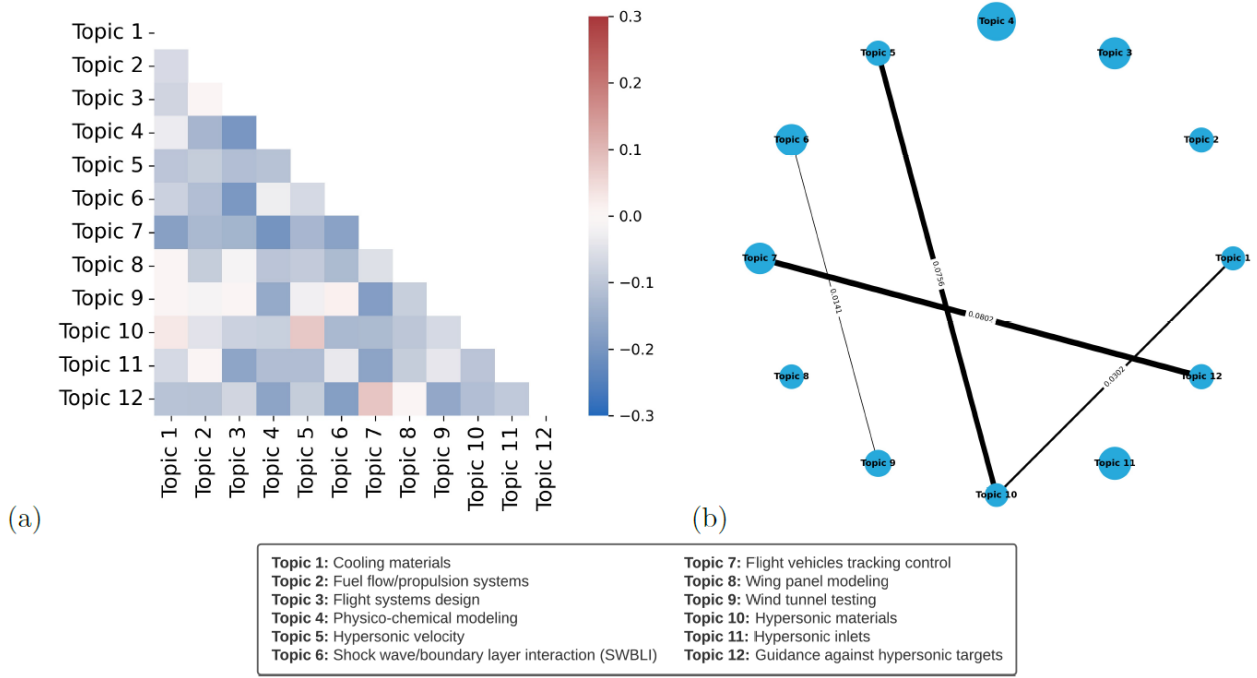

**Figure 8.** Topics correlations, **(a)** correlation heatmap, and **(b)** positive correlations graph.

*Evolution of the research themes*

After building the STM model and extracting the main research topics and in order to analyze the evolution of topics over time, the proportion of each publication was regressed on the date of publication. More specifically, the conditional expectation of topic prevalence given the characteristics of the publications and their date of publication was estimated. Figure 9-a shows the results. Shaded areas in the figure represent the 95% confidence interval. As seen in the figure, 6 topics, i.e., "Topic 1", "Topic 6", "Topic 7", "Topic 8", "Topic 11", and "Topic 12", followed an increasing trend over time and the other 6 topics' prevalence decreased. Among topics with an increasing trend, "Topic 7", i.e., flight vehicles tracking control, and "Topic 12", i.e., guidance against hypersonic targets, had the steepest increase. On the other hand, "Topic 3", i.e., flight systems design, and "Topic 4", i.e., physico-chemical modeling, had the sharpest decrease among topics with a decreasing trend. The other observation is that at the beginning of the period, researchers' focus was mostly on "Topic 3" and "Topic 4", however, the attention was shifted more to "Topic 7" and "Topic 12" in the final period. In addition, from the figure, it is seen that researchers' focus on "Topic 11", i.e., hypersonic inlets, was almost constant over time. "Topic 7", "Topic 12", and "Topic 4" are the top-3 most prevalent topics in 2020.

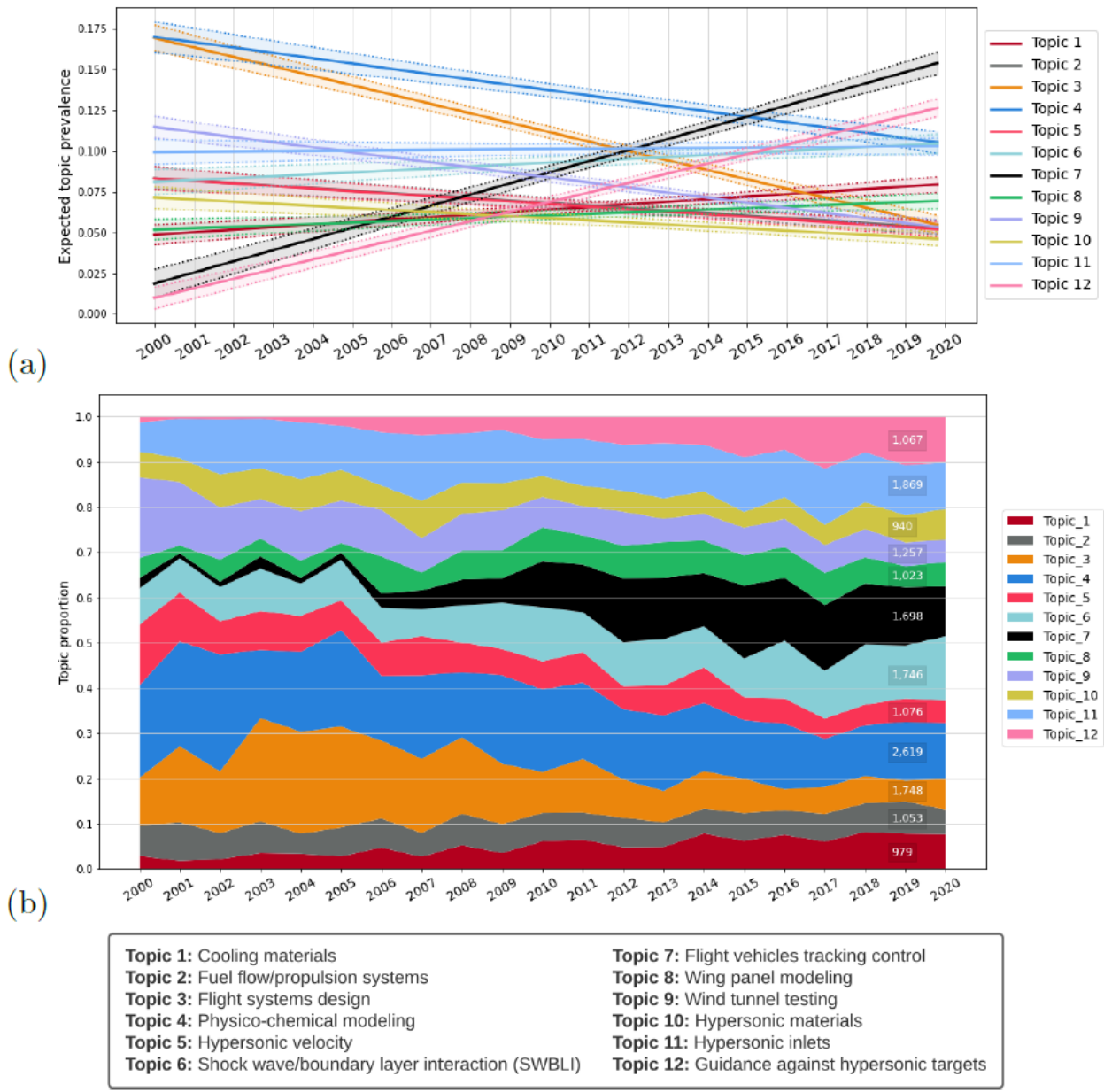

**Figure 9.** Temporal evolution of hypersonics research topics, **(a)** topic prevalence in publications from 2000 to 2020. The shaded areas between the dotted lines indicate the 95% confidence interval, and **(b)** dominant topic distribution across publications over time. The numbers on the figure represent the total number of publications dominated by the respective topic.

Figure 9-b shows the results for the distribution of dominant topics across publications over time. In STM, each publication can cover more than one topic as topics are assigned to publications with a probability. To calculate the distribution of dominant topics across publications, the publication-topic probability matrix was considered. For each publication, only the topic with the highest probability was assigned. As seen in the figure, the results are in line with our previous observations in Figure 9-a. "Topic 3" and "Topic 4" were the most dominating topics in the beginning years. Although "Topic 4" proportion has decreased over time, it still remains one of the most dominating topics in the final period in terms of publication covered, placing it among the top-3 dominating topics in 2020 next to "Topic 6", i.e., SWBLI, and "Topic 7". Among these topics, "Topic 7" was the one with the lowest proportion in the beginning years. Similarly, the attention to "Topic 12" has increased drastically over time.

**Discussion and conclusion**

The field of hypersonics is wide, and complex, and has been witnessing tremendous developments in recent years. With defence and commercial applications, it is said that the technology has the potential to transform the industry (Deloitte 2020). In this study, we focused on hypersonic-related scientific publications within the period from 2000 to 2020, and employed machine learning and NLP techniques to characterize the research landscape and its evolution over time. We measured publication similarity over time and identified some patterns that are indicative of cycles during two decades of research. We also used structural topic modelling to identify 12 key topics in hypersonics research. The identified topics offer comprehensive and logical coverage of the research field and are relatively similar to the topics being used to structure review papers (Sziroczak and Smith 2016) or scientometric studies (Senay 2017) around hypersonics. The fact that topics are algorithmically extracted removes subjectivity from the exercise and enables consistent comparisons between topics and between time intervals.

The number of publications has been growing in almost all scientific disciplines. It was observed that the field of hypersonics has been no exception with publications increasing over time. However, our findings suggest that the focus on research topics has been shifting gradually, due to the dynamic nature of the science and in order to reflect changing needs. Based on the results, although the focus was mainly on the design of hypersonic flight systems and physico-chemical modelling in hypersonic flow simulation, in the beginning, non-linear adaptive tracking control of hypersonic flight vehicles and robust guidance against hypersonic targets attracted more attention in the final year. Moreover, research on tracking control of hypersonic flight vehicles has experienced the sharpest increase over the examined period, illustrating one of the challenges of hypersonic flight. The importance of optimal performance for hypersonic flight control is also reflected in the literature. For instance, a line of research is investigating a shift from basic control performance including, e.g., stability, and robustness, to the design of fuzzy optimal tracking controllers for HFVs (Bu and Qi 2020).

Aerothermodynamics, propulsion, and structures are identified as the three major challenge areas associated with hypersonic vehicles (Sziroczak and Smith 2016). These research areas are reflected well in our extracted topics. In addition to these lines of research, our findings also demonstrate an emphasis on modelling and simulation research projects. Of note, recent developments in computer science, e.g., deep learning techniques, have opened up new directions for hypersonic vehicles' trajectory simulation and optimization (e.g., Shi and Wang 2020, 2021). As another example, such powerful machine learning techniques have been also used for hypersonic flight control recently (Xu et al. 2014), a category that is represented well in our extracted topics. In fact, flight control systems have been influenced by artificial neural networks and rapidly evolved during the last two decades (Emami et al. 2022).

The fact that most steps of the workflow can be automated is significant. It means that insights can be generated quickly, even for complex research fields involving thousands of papers every year, to assist analysts and decision-makers in better understanding the research dynamics and help with research and development (R&D) strategies. This is of particular importance for R&D that requires fast progress, e.g., COVID-19 research (Ebadi et al. 2021), or disruptive technology development potentially affecting strategic stability (Sechser et al. 2019), economic development (Rifkin 2011), and national security. With defence and commercial applications, hypersonics research will continue to bring innovation to aeronautic and space industries (Deloitte 2020), but other examples of such disruptive technologies abound, e.g., deep learning (LeCun et al. 2015), quantum computing (Gyongyosi and Imre 2019), mRNA vaccines (Sahin et al. 2014). The nations and organisations able to best understand and monitor the research landscape around them will have a competitive advantage.

**Limitations and future work**

We used scientific publications in the period from 2000 to 2020 to characterize the landscape of hypersonics research. Other data sources, e.g., patents, and time intervals could be used to perform complementary investigations. It should be noted that our findings may only reflect the focus of researchers in the field of hypersonics at a very high level. Future works may consider other levels of abstraction using our proposed methodology. Temporal fusion and/or division of research themes can be also investigated in future research. In addition, this work was based on analysis of uni-grams (tokens) and could benefit from an n-gram approach in the future. Finally, we performed the analysis using titles and abstracts of publications as we did not have access to the full text of the collected publications. Future work may consider the full body of publications.